\documentclass{IEEEtran}

\usepackage{pgfplots}
\pgfplotsset{compat=1.5}
\pgfplotsset{every axis/.append style={thick}}
\selectcolormodel{gray} 

\usepackage{cite} 
\usepackage{multirow}
\usepackage{booktabs}

\usepackage{amsmath,amsfonts}
\usepackage{enumerate} 

\usepackage{algorithm} 
\usepackage{algorithmic}  

\usepackage{tabularx}
\usepackage{lipsum}

\newtheorem{lemma}{Lemma} 
\newtheorem{theorem}{Theorem} 
\newtheorem{definition}{Definition} 
   
\newtheorem{corollary}{Corollary}

\newcommand*{\defeq }{\stackrel{\text{def}}{=}}  

\newcommand{\opt}{\mathrm{opt}}
\newcommand{\non}{\mathrm{non}}

\begin{document}

\title{An Analytic Expression of Relative Approximation Error   for a Class of Evolutionary Algorithms}
\author{Jun He 
\thanks{This work was supported by EPSRC under Grant No. EP/I009809/1}
\thanks{Jun He is with Department of Computer Science,  Aberystwyth University, Aberystwyth SY23 3DB,    United Kingdom, Email:  jun.he@aber.ac.uk } }

\maketitle

\begin{abstract} 
An important   question in evolutionary computation is how good
solutions evolutionary algorithms can produce. This paper aims to provide an
analytic analysis of solution quality in terms of the
relative approximation error, which is defined by   the error between 1 and the
approximation ratio of the  solution found by an evolutionary algorithm. Since evolutionary algorithms are iterative methods, the
relative approximation error  is  a function of generations. With the help of matrix
analysis,   it is possible   to obtain an exact expression of such a
function.  In this paper,  an analytic expression for calculating the
relative approximation error    is presented  for  a class of evolutionary
algorithms, that is, (1+1) strictly elitist evolution algorithms.  Furthermore,
analytic expressions of the fitness value and the average convergence
rate in each generation are also derived for this class of evolutionary algorithms. The approach is promising, and it can be extended to non-elitist or population-based algorithms too.
\end{abstract}

\section{Introduction}
\label{sec:introduction}
Evolutionary algorithms (EAs) have been widely used to find good  solutions to hard   optimization problems. Many experimental results  claim that EAs can obtain good quality solutions quickly. Nevertheless, from the viewpoint of the $NP$-hard theory,   no efficient algorithm exists for solving NP-hard combinatorial optimization problems at the present and possibly for ever. Therefore it is unlikely that EAs are  efficient  in solving hard combinatorial optimization problems too.  Instead of searching for the exact solution to hard optimization problems, it is  more reasonable to expect that EAs are able to find some good approximate solutions efficiently.

It is necessary to answer the question of how good  solutions EAs can produce to hard  optimization problems in terms of the approximation ratio.   Current work focuses on the approximation ratio of the solution found by an EA within polynomial time.  The research has attracted a lot of interests in recent years. Various combinatorial optimization problems have been investigated, including the  minimum vertex cover problem~\cite{oliveto2009analysis,friedrich2010approximating}, the partition problem~\cite{witt2005worst},   the set cover problems  \cite{yu2012approximation}, the minimum label spanning tree problem~\cite{lai2014performance},  and many others.

This paper studies the approximation ratio of EAs from a different viewpoint. It aims to estimate the relative approximation error  of the best solution found by an EA in each generation, but without considering whether the EA is an approximation algorithm or not. The problem in this paper is described as follows: Given an EA for maximizing a fitness function $f(x)$, let  $f_{\opt}$ be the optimal fitness and $F_t$ the expected fitness value of the  best solution    found in the $t$th generation. The approximation ratio of the $t$th generation solution   is 
$F_t/f_{\opt}$. The approximation ratio  of the optimal solution is  1.  The   relative approximation error is 
\begin{equation}
E_t= 1-\frac{F_t}{f_{\opt}}.
\end{equation}  
In~\cite{he2003analysis}, $E_t$ is called the performance ratio. In order to avoid   confusion with the approximation ratio, it is renamed the relative approximation error. 
 The relative approximation error  $E_t$ is a function of $t$.
Our main  research question is to find an upper bound $\beta(t)$ on the error $E_t$. 

 The perfect answer is to obtain a  function $\beta(t)$ in a closed form such that $E_t=\beta(t)$. For (1+1) strictly elitist EAs,   such an analytic expression   has been constructed  in this paper using matrix analysis. To the  best of our knowledge, this is the first result of   expressing  the relative approximation error  (also the fitness value and the average  convergence rate) in a closed form for a class of EAs.

The paper is arranged as follows: Section~\ref{secLinks} reviews the links to related work.  Section~\ref{secApproximation} defines the relative approximation error. Sections~\ref{secExample} and~\ref{secExamplec} conduct a case study.  Section~\ref{secMarkov} introduces Markov modelling. Section~\ref{secAnalysis} makes a  theoretical analysis.    Section~\ref{secConclusions} summarizes the paper.

\section{Links to related work}
\label{secLinks}
The  relative approximation error   belongs to   the  convergence rate study of EAs, which can be traced back to 1990s~\cite{suzuki1995markov,suzuki1998further,he1999convergence}.  This paper   only investigates  EAs for discrete optimisation, although the converegnce rate of EAs for  continuous optimization~\cite{rudolph1997local,rudolph1997convergence,rudolph2013convergence} is also important.    EAs belong to  iterative methods. A fundamental question in iterative methods is the  convergence rate, which can be formalised as follow~\cite{he1999convergence}.  Since the $t$th generation solution is a random variable, we let $\mathbf{p}_t$ be a vector representing  its probability distribution over the search space, $\boldsymbol{\pi}$  a vector such that $\pi(\mathcal{S}_{\opt})=1$ for the optimal solution set $\mathcal{S}_{\opt}$ and $\pi(\mathcal{S}_{\non})=0$ for the non-optimal solution set $\mathcal{S}_{\non}$. The  convergence rate problem asks the question how fast  $\mathbf{p}_t$ converges to $\boldsymbol{\pi}$. The goal is to obtain a bound $\beta(t)$ such that $\parallel \mathbf{p}_t-  \boldsymbol{\pi} \parallel \le \beta(t)$, where $\parallel \cdot \parallel$ is a norm.    There are various ways to assign the norm.  For example, if the $t$th generation solution is a binary string $x_1 \cdots x_n$ and  the optimal solution  is  $1 \cdots 1$, the norm is set to be the Hamming distance:
\begin{align}
\label{equNorm1}
\parallel \mathbf{p}_t-  \boldsymbol{\pi} \parallel=   \sum_i |1-x_i|.
\end{align} 
 In the current paper, the norm is set to be the relative approximation error $\parallel \mathbf{p}_t-  \boldsymbol{\pi} \parallel = 1-F_t/f_{\opt}$.

 According to~\cite{ming2006convergence}, there are two approaches to analyse the convergence rate of EAs for discrete optimization.  The first approach is based on the eigenvalues of the transition submatrix associated with an EA.   Suzuki~\cite{suzuki1995markov}   derived a lower
bound of convergence rate for   simple genetic algorithms through analysing eigenvalues   of the  transition matrix.   Schmitt and Rothlauf~\cite{schmitt2001importance} found that the
convergence rate   is determined by the second largest
eigenvalue of the transition matrix.   The approach used in the current paper is the same as that in~\cite{suzuki1995markov,schmitt2001importance}. All are based on analysing the powers and eigenvalues of the transition matrix. The other approach is
based on  Doeblin's condition~\cite{he1999convergence,ding2001convergence}.   Using the minorisation condition in   Markov chain theory, He and Kang~\cite{he1999convergence} proved that   for the
EAs with time-invariant genetic operators, the convergence rate can be upper-bounded by
$\epsilon^{t}$ where  $\epsilon \in (0,1)$.    

The research in this paper is also linked to fixed budge analysis. Jansen and Zarges~\cite{jansen2012fixed,jansen2014performance} proposed fixed budget analysis. It aims to find  lower and upper bounds  $\beta_{low}(b)$ and $\beta_{up}(b)$ such that  $\beta_{low}(b)  \le f_b \le \beta_{up}(b)$ usually for  a fixed budget $b$.    They investigated  two algorithms: random local search and the $(1+1)$ EA and obtained such bounds on several pseudo-Boolean optimization problems. 

The convergence rate study can implement the same task as fixed budget analysis does. Provided that  $ 1 -f_b/f_{\opt} \le \beta(b)$, it is trivial to derive  $ f_b \ge f_{\opt} (1-\beta(b))$. 
Nevertheless,  there are   significant differences between the convergence rate study and fixed budget analysis. 
\begin{itemize}
\item The  convergence rate study   has existed in EAs for more than two decades~\cite{suzuki1995markov,suzuki1998further,he1999convergence}.   Fixed budget analysis was recently proposed by Jansen and Zarges~\cite{jansen2014performance}.

\item The  convergence rate study    focuses on estimating the error  $\parallel \mathbf{p}_t-  \boldsymbol{\pi} \parallel$, where  the norm $\parallel \cdot \parallel$ can be chosen as the absolute   error $f_{\opt}-F_t$, relative  error $1-F_t/f_{\opt}$ or Hamming distance. Fixed budget analysis   aims  at bounding $f_b$  for a fixed budget  $b$ (that is a fixed number of generations)~\cite{jansen2014performance}.   

\item   In   the convergence rate study,  the upper bound $\beta(t)$ on  $\parallel \mathbf{p}_t-  \boldsymbol{\pi} \parallel $ usually is an exponential function or combination of linear functions of $t$~\cite{he1999convergence}.    In fixed budget analysis, the bound   $\beta (b) $ on $f_b$ may not be an exponential function of $b$~\cite{jansen2014performance}.  

\item In the convergence rate,   the bound   on  $\parallel \mathbf{p}_t-  \boldsymbol{\pi} \parallel $ holds for all $t$. But in fixed budget analysis,  the bound  on $f_b$ often is estimated for a fixed budget $b$~\cite{jansen2014performance}. 

\item Matrix analysis is widely used in the convergence rate study~\cite{suzuki1995markov,schmitt2001importance}, but it is not used in fixed budget analysis.
 
\end{itemize}

\section{Relative approximation error, fitness value and average convergence rate}
\label{secApproximation} 
Consider a maximization problem, that is,   $\max \{ f(x); x \in \mathcal{S}\}$ where $\mathcal{S}$ is a finite set,  and $f_{\opt} > f(x) \ge 0$. For the sake of analysis, let $\mathcal{S}=\{ 0, 1, \cdots, L\}$ denote the set of all solutions.  We  assume that
$$ f_{\max} =f(0)  > f(1) \ge \cdots \ge f(L) = f_{\min}.$$ 
$\mathcal{S}$ is split into two subsets: the optimal solution set $\mathcal{S}_{\opt}=\{0\}$ and the set of non-optimal solutions $\mathcal{S}_{\non} =\{1, \cdots, L\}$.

An  EA  for solving the above  problem is regarded as an iterative procedure:  initially construct a population of solutions $\Phi_0$;    then given the $t$th generation population  $\Phi_t$, generate a new population  $\Phi_{t+1}$ in a probabilistic way. This procedure is repeated until an optimal solution is found. This paper investigates a class of (1+1) elitist EAs which are  described in Algorithm~\ref{alg1}. This kind of EAs is very popular in the theoretical analysis of EAs.

\begin{algorithm}
\caption{A (1+1) Strictly Elitist EA} \label{alg1}
\begin{algorithmic}[1]
\STATE set $t \leftarrow 0$;
\STATE $ \Phi_{0}\leftarrow$  choose a solution from $\mathcal{S}=\{ 0, 1, \cdots, L\}$;
 
\WHILE{$\Phi_t$ is not an optimal solution}
\STATE $\Psi_{t} \leftarrow$ mutate $\Phi_t$;
\IF{$f(\Psi_t) > f(\Phi_t)$}
\STATE $\Phi_{t+1} \leftarrow$ select $ \Psi_{t}$;
\ELSE
\STATE  $\Phi_{t+1} \leftarrow $ select $\Phi_{t}$;
\ENDIF
\STATE   $t\leftarrow t+1$;
\ENDWHILE
\end{algorithmic}
\end{algorithm}

 The expression $\Phi_t=x$ means  the $t$th generation individual $\Phi_t$ at state $x$ where $\Phi_t$ is a random variable and $x$ its value taken from $\mathcal{S}$. The fitness of   $\Phi_t$ is denoted by $f(\Phi_t)$. Since $f(\Phi_t)$ is a random variable, we consider its expectation $
F_t \defeq \mathrm{E}[f(\Phi_t)].
$ 
 The approximation ratio of the $t$th generation individual is 
${F_t}/{f_{\opt}}.$ The approximation ratio of the optimal solution is $1$.
\begin{definition}
The relative approximation error  of the $t$th generation individual is defined by
\begin{equation}
E_t = 1- \frac{F_t}{f_{\opt}}.
\end{equation}   
\end{definition}

There is a  link between the relative approximation error  and  the fitness value. From the definition of the relative approximation error,  
we know that 
the fitness value in the $t$th generation equals to
\begin{align}
F_t =f_{\opt}(1-E_t).
\end{align} 

There is a  link between the relative approximation error  and average convergence rate~\cite{he2016average}.
From the definition of  the (geometric) average  of the  convergence rate    of an EA for $t$ generations, we get 
\begin{align}
\label{equAverageRate}
   R_t&\defeq  1- \left( \left| \frac{ f_{\opt}- F_t}{ f_{\opt}- f_0} \right|\right)^{1/t} =1-\left( \frac{E_t}{E_0}\right)^{1/t}.
\end{align}

\section{Example}
\label{secExample} 
Given $F_t$, $E_t$ and $R_t$, which  is the best option to measure the performance of an EA? We use a simple experiment to show their advantage and disadvantage.     Consider  the problem of maximizing a pseudo-Boolean  function $f(x)$ where $x=x_1 \cdots x_n$ is a binary string. Three test functions are used in the experiment. 
  \begin{align*}
  \mbox{OneMax function }
 &f_{\mathrm{one}}(x)= |x|,
\\
  \mbox{square function } &f_{\mathrm{squ}}(x)=|x|^2,\\
\mbox{logorithmic function } &f_{\mathrm{log}}(x)= \ln (|x|+1), 
\end{align*}     
where   $|x| = x_1+ \cdots +x_n$. A (1+1)  EA is used for solving the optimisation problem. This EA is also called randomised local search.
\begin{algorithmic} 
\STATE \textbf{Onebit Mutation.} Given a binary string, chose one bit at random and then flip it.
\STATE \textbf{Elitist Selection.} Choose the best from  the parent and child as the next parent.
\end{algorithmic}

Onebit mutation is chosen for the sake of demonstrating that the average convergence rate $R_t$ may equal to a constant, according to the theory of the average convergence rate~\cite{he2016average}.  
The three functions are the easiest  to the (1+1) EA among all pseudo-Boolean functions whose  optimum is unique at $1 \cdots 1$ according to the theory of the easiest and hardest functions~\cite{he2015easiest}.   

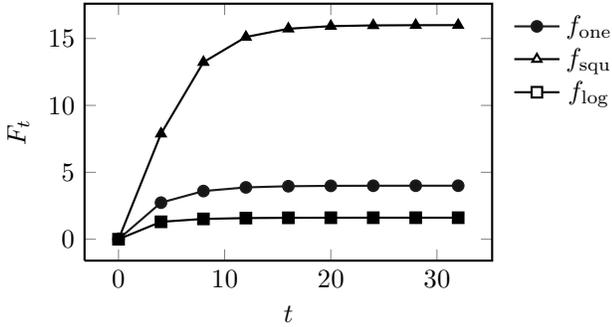
\begin{figure}[ht]
\begin{center}
\begin{tikzpicture}
\begin{axis}[domain=0:35, width=7cm,height=5cm,
scaled ticks=false,
legend style={draw=none},legend pos= outer north east,
xlabel={$t$}, 
ylabel={$F_t$},    
]  
\addplot table[only marks]   {F-f1-0.dat};  
  \addplot[mark=triangle*] table[ only marks]   {F-f2-0.dat};  
    \addplot[mark=square*] table[  only marks]   {F-f3-0.dat};

    \legend{$f_{\mathrm{one}}$, $f_{\mathrm{squ}}$, $f_{\mathrm{log}}$ ,}
        \end{axis}
\end{tikzpicture} 
\caption{Fitness value $F_t$.}
\label{fig-1}
\end{center}
\end{figure}

In the experiment, we set $n=4$. This small value is chosen for the sake of displaying  matrices   in Section~\ref{secExamplec} in one column. The initial solution is set to $0000$. We  run the EA $10^8$ times.  The EA stops after 35 generations for each run.
Fig.~\ref{fig-1} demonstrates the fitness value $F_t$  which is averaged over $10^8$ runs. The figure  shows that $F_t$ converges to $4$  on   $f_{\mathrm{one}}$,     $16$ on   $f_{\mathrm{squ}}$ and  $\ln 5$ on  $f_{\mathrm{log}}$. But it is not clear how $F_t$ is close to $f_{\opt}$, and   how fast $F_t$ converges to $f_{\opt}$.

\begin{figure}[ht]
\begin{center}
\begin{tikzpicture}
\begin{axis}[domain=0:35, width=7cm,height=5cm,
scaled ticks=false,
legend style={draw=none},legend pos=outer north east,
xlabel={$t$}, 
ylabel={$E_t$}, 
]  
\addplot table[only marks]   {ARE-f1-0.dat};  
\addplot[mark=triangle*] table[ only marks]   {ARE-f2-0.dat};  
\addplot[mark=square*] table[  only marks]   {ARE-f3-0.dat};  
\legend{$f_{\mathrm{one}}$, $f_{\mathrm{squ}}$, $f_{\mathrm{log}}$, $f_{\mathrm{one}}$, $f_{\mathrm{squ}}$, $f_{\mathrm{log}}$,}
\end{axis}
\end{tikzpicture} 
\caption{Relative approximation error  $E_t$.}
\label{fig-2}
\end{center}
\end{figure}
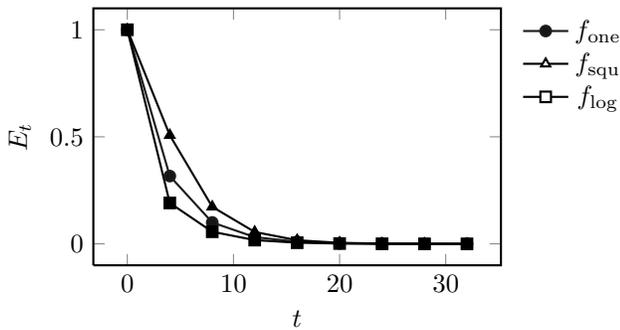

Fig.~\ref{fig-2} presents the relative approximation error  $E_t$, which converges to 0. From the figure, we observe that for any $t$,  $E_t$ on   $f_{\mathrm{log}}$ is smaller than that on $f_{\mathrm{one}}$, then smaller than that on $f_{\mathrm{squ}}$.    

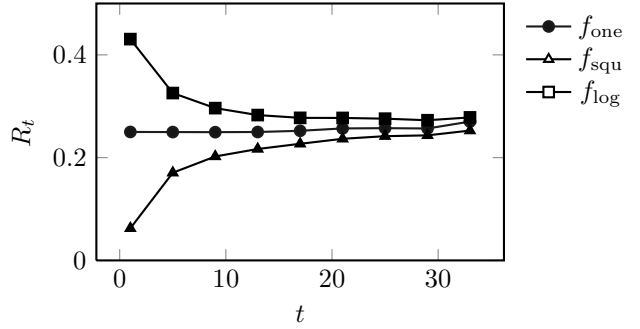
\begin{figure}[ht]
\begin{center}
\begin{tikzpicture}
\begin{axis}[domain=1:35, width=7cm,height=5cm,
scaled ticks=false,
legend style={draw=none},legend pos= outer north east,
xlabel={$t$}, 
ylabel={$R_t$}, 
ymin=0,
ymax=0.5,  
]  
 \addplot table[only marks]   {ACR-f1-0.dat};  
  \addplot[mark=triangle*] table[ only marks]   {ACR-f2-0.dat};  
    \addplot[mark=square*] table[  only marks]   {ACR-f3-0.dat};   
    \legend{$f_{\mathrm{one}}$, $f_{\mathrm{squ}}$, $f_{\mathrm{log}}$}
        \end{axis}
\end{tikzpicture} 
\caption{Average convergence rate $R_t$.}
\label{fig-3}
\end{center}
\end{figure}

Fig.~\ref{fig-3} illustrates the  average convergence rates  $R_t$,   which converges  to   $0.25$. 　 From the figure, we see the   difference of the average convergence rate on the three functions.
\begin{itemize}
\item    $ R_t= 0.25$ on $f_{\mathrm{one}}$. The EA  converges    as fast as an exponential decay: 
  $
E_t =0.75^t E_0$. 

\item  $ R_t$ converges to $0.25$ on $f_{\mathrm{squ}}$ but its value is larger than $0.25$ . The EA  converges faster than the exponential decay: $
E_t \le 0.75^t E_0$. 

\item   $ R_t$ converges to $0.25$ on $f_{\mathrm{log}}$ but its value is smaller than  $0.25$ . The EA  converges slower than the exponential decay: $
E_t \ge 0.75^t E_0$. 

\end{itemize}

\section{Markov chain modelling for (1+1) strictly elitist EAs}
\label{secMarkov}
This section introduces Markov chain modelling for (1+1) strictly elitist EAs. It follows the Markov chain framework described in~\cite{he2003towards,he2016average}.

Genetic operators in EAs can be either time-invariant or time-variant~\cite{rudolph1998finite,he1999convergence}. This paper only considers time-invariant operators. Such an EA  can be modelled by a homogeneous Markov chain with  transition probabilities 
$$
r_{i,j}\defeq\Pr(\Phi_{t+1}= i \mid  \Phi_t = j),  \quad  i , j \in \mathcal{S}.
$$

According to the strictly elitist selection, transition probabilities satisfy
\begin{equation}
r_{i,j}=
\left
\{\begin{array}{lll}
\ge 0, &\mbox{if } f(i)> f(j),\\
\ge 0, &\mbox{if } i=j,\\
=0, &\mbox{otherwise}.
\end{array}
\right.
\end{equation} 

Let   $\mathbf{R} $  denote the transition submatrix which represents transition  probabilities  among non-optimal states $\{1, \cdots, L \}$. It is a $L \times L$ matrix, given as follows:
\begin{align} 
\label{equMartixR}
\mathbf{R} =  \begin{pmatrix}  
  r_{1,1}    & r_{1,2}& r_{1,3}& \cdots &   r_{1,L-1}  &r_{1,L}  \\ 
   0& r_{2,2} & r_{2,3}& \cdots & r_{2,L-1} &r_{2,L}\\
0 &  0 & r_{3,3} &  \cdots & r_{3,L-1} &r_{3,L}\\
 \vdots & \vdots &  \vdots  & \vdots & \vdots    \\
 0&  0&  0&\cdots & 0& r_{L,L}     \\
 \end{pmatrix}.
\end{align}

  Let $p_t(i) = \Pr(\Phi_t=i)$ denote the probability  of  $\Phi_t$  at state  $i$ and the vector 
$$
\mathbf{q}_t\defeq (p_t(1), p_t(2), \cdots, p_t(L) )^T.
$$ 
 Here notation  $\mathbf{v}$ is a column vector and $\mathbf{v}^T$ the row   with the transpose operation.

For any $t \ge 1$, the probability  $p_t(i)$ (where $i \in \mathcal{S}_{\non}$)  equals to  
\begin{align*}
\Pr(\Phi_{t}=i)=&\sum_{j\in S_{\non}} \Pr(\Phi_{t}= i \mid  \Phi_t = j) \Pr(\Phi_{t-1}=j)\\
=&\sum_{i \in S_{\non}} p_{t-1}(i) r_{i,j}. 
\end{align*}
It can be represented by   matrix iteration
\begin{align}
&\mathbf{q}_{t}= \mathbf{R} \mathbf{q}_{t-1}    = \mathbf{R}^t \mathbf{q}_0 . 
\label{equMatrixIteration}  
\end{align}

Let  $e(i)=f_{\opt}-f(i)$  denote  the fitness error between  the optimal solution and   each non-optimal solution and the vector   　
\begin{align*}
\mathbf{e}^T \defeq  
 (e(1), e(2), \cdots, e(L) ) .
\end{align*} 
 
Then  the relative approximation error $E_t$ can be represented by 
\begin{align}
\label{equARE} 
 E_t= \frac{  \mathbf{e}^T \mathbf{q}_t }{f_{\opt}} 
 = \frac{ \mathbf{e}^T \mathbf{R}^t  \mathbf{q}_0   }{f_{\opt}}.
\end{align}  

From formula~(\ref{equARE}), we see that $E_t$ is determined by the initial distribution $\mathbf{q}_0$,  matrix power   $\mathbf{R}^t$,   fitness error $\mathbf{e}^T$ and   optimal fitness value $f_{\opt}$. Only $\mathbf{R}^t$ is a function of   $t$, so it plays the most important role in determining the relative approximation error.

\section{An analytic expression of relative approximation error}
\label{secAnalysis} 
This section gives an analytic expression of the relative approximation error  for (1+1) strictly elitist EAs. The analysis is based on an existing  result in matrix analysis~\cite{shur2011simple,huang1978efficient}.

From  (\ref{equARE}), we see that calculating $E_t$ becomes a mathematical problem of expressing the  matrix power $\mathbf{R}^t$ once the initial probability distribution $\mathbf{q}_0$ and the fitness error $\mathbf{e}^T$ are known.
For (1+1) strictly elitist EAs,  the matrix $\mathbf{R}$ is an upper triangular, and then it is feasible to  express matrix $\mathbf{R}^t$ explicitly in terms of its entries  in a closed form\cite{shur2011simple}. 

For the sake of simplicity, matrix  $\mathbf{R}$ is assumed to satisfy the following condition:
\begin{itemize}
\item  \textbf{Unique condition:} transition probabilities $r_{i,i} \neq r_{j,j}$ if $i \neq j$.
\end{itemize}     If transition probabilities $r_{i,i} =r_{j,j}$  for some $i \neq j$, a similar discussion can be conducted  but will be given in a separate paper. 

\begin{definition}
 The power factors of $\mathbf{R}$, $[p_{i,j,k}]$ (where $i,j,k=1, \cdots, L$), are recursively defined as follows:
\begin{align} 
p_{j,j,j}=& r_{j,j},  
\label{equDef1}\\
p_{i,j,k}=&0,                &k<i \mbox{ or } k >j,
\label{equDef2}\\
p_{i,j,k} =&\frac{\sum^{j-1}_{l=k} p_{i,l,k} r_{l,j}}{r_{k,k}-r_{j,j}}, &i \le k< j,
\label{equDef3}\\
p_{i,j,j} =&r_{i,j}-\sum^{j-1}_{l=i} p_{i,j,l}, &i<j.
\label{equDef4}
\end{align}
\label{defMain} 
\end{definition}

Lemmas \ref{lemMain} and \ref{lemPower} show  how to calculate  the matrix power $\mathbf{R}^t$. For the sake of completeness, their proofs~\cite{shur2011simple} are given here. 
\begin{lemma} [Lemma~1.2 in~\cite{shur2011simple}]
Let $\mathbf{R}= [r_{i,j}]$ be a non-singular upper triangular matrix with unique diagonal entries.
Denote the entries of the  matrix power $\mathbf{R}^t$ by $[r_{i,j|t}]$.  For any $ t \ge 1$, if $r_{i,j|t}= \sum^j_{k=i} p_{i,j,k} (r_{k,k})^{t-1}$, then $r_{  i,j|t+1}=\sum^j_{k=i} p_{i,j,k} (r_{k,k})^t$.
\label{lemMain}
\end{lemma}

\begin{IEEEproof} 
Since $\mathbf{R}^{t+1}=\mathbf{R}^t  \cdot \mathbf{R}$,    $r_{i,l|t}=0$ if $l <i$ (because $\mathbf{R}^t$ is upper triangular) and $r_{l,j}=0$ if $l >j$ (because $\mathbf{R}^t$ is upper triangular), we have
\begin{align}
r_{i,j |t+1} = \sum^j_{l=i} r_{i,l |t} r_{l,j}. 
\end{align}

From the assumption:   $r_{i,j|t}= \sum^j_{k=i} p_{i,j,k} (r_{k,k})^{t-1}$, and noting that $p_{i,l,k}=0$ if $k >l$, we have
\begin{align}
r_{i,j |t+1} =&\sum^j_{l=i} r_{l,j}  \sum^l_{k=i} p_{i,l,k} (r_{k,k})^{t-1} 
\nonumber \\
=&\sum^j_{k=i} (r_{k,k})^{t-1} \sum^j_{l=k} r_{l,j} p_{i,l,k}.
\label{equSumSum}
\end{align}

Notice that 
\begin{align}
\sum^j_{l=k} r_{l,j} p_{i,l,k} = \sum^{j-1}_{l=k} r_{l,j} p_{i,l,k}+ r_{j,j}p_{i,j,k}.
\label{equSplitSum}
\end{align}
Then substituting the sum in (\ref{equSplitSum}) by (\ref{equDef3}) in Definition~\ref{defMain}, we have
\begin{align}
\sum^j_{l=k} r_{l,j} p_{i,l,k} &=  p_{i,j,k}(r_{k,k}-r_{j,j})+ r_{j,j}p_{i,j,k}
\nonumber \\
&=p_{i,j,k} r_{k,k}.
\end{align}

Finally (\ref{equSumSum}) is simplified as $r_{i,j | t+1} =\sum^j_{k=i} p_{i,j,k} (r_{k,k})^t$. This is the required conclusion.
\end{IEEEproof}

\begin{lemma}[Theorem~1.3 in~\cite{shur2011simple}]
Let $\mathbf{R}= [r_{i,j}]$ be a non-singular upper triangular matrix with unique diagonal entries. For any $t \ge 0$,  
\begin{align}
&r_{i,j|t+1} =\sum^j_{k=i} p_{i,j,k} (r_{k,k})^{t-1} =\sum^j_{k=i} p_{i,j,k} (r_{k,k})^{t}.  \label{equPower} 
\end{align}
\label{lemPower}
\end{lemma}

\begin{IEEEproof}
According to (\ref{equDef1}), (\ref{equDef2}) and (\ref{equDef4}) in Definition~\ref{defMain},   we see that (\ref{equPower}) is true for $t=1$. Then by induction, (\ref{equPower}) is true for all $t >1$ from Lemma~\ref{lemMain}.
\end{IEEEproof}

The above lemma gives an analytic expression of the matrix power $\mathbf{R}^t$.  Given a $L\times L$ matrix $\mathbf{R}$,  the time complexity of   calculating $\mathbf{R}^t$  is $2 \binom{L+2}{3} + L(t-3)$ in terms of the number of multiplication and divisions~\cite{shur2011simple}.

For the sake of notation, $e(i)$  is  denoted by $e_i$  and $q_0(i)$ by $q_i$. Define coefficients
\begin{align}
\label{equAlpha}
&c_k\defeq \frac{   \sum^L_{i=1} \sum^L_{j=i} e_i    p_{i,j,k}  q_j }{f_{\opt}},  &k=1, \cdots, L,
\end{align}
where  $c_k$ is independent of $t$.

\begin{theorem}
\label{theMain}
If $\mathbf{R}= [r_{i,j}]$ is a non-singular upper triangular matrix with unique diagonal entries, then for any $t \ge 1$, the relative approximation error $E_t$ is expressed by
\begin{align}
\label{equARE3} 
 E_t =\sum^{L}_{k=1}  c_k \lambda_k^{t-1},
\end{align}
where $\lambda_k=r_{k,k}$  are eigenvalues of matrix $\mathbf{R}$. 
\end{theorem}
\begin{IEEEproof}
From (\ref{equARE}), we know
\begin{align} 
\label{equARE2}
 E_t =&\frac{ \mathbf{e}^T \mathbf{R}^t    \mathbf{q}_0 }{f_{\opt}}=\frac{  \mathbf{e}^T\mathbf{R}^t    \mathbf{q}_{0}}{f_{\opt}}.
\end{align}   
Using (\ref{equPower}),   we get 
\begin{align} 
 \mathbf{e}^T\mathbf{R}^t    \mathbf{q}_{0} =\sum^L_{i=1}  \sum^L_{j=1}   \sum^j_{k=i} e_i p_{i,j,k} (r_{k,k})^{t-1} q_j.
\end{align} 

According to Definition~\ref{defMain}, $p_{i,j,k}=0$ if $k<i$ or $k>j$ and $p_{i,j,k}=0$ if $i>j$, then 
\begin{align}
 \mathbf{e}^T\mathbf{R}^t    \mathbf{q}_{0} &=\sum^L_{i=1}  \sum^L_{j=1}   \sum^L_{k=1} e_i p_{i,j,k} (r_{k,k})^{t-1} q_j\nonumber \\
 &= \sum^{L}_{k=1} (r_{k,k})^{t-1}  \sum^L_{i=1} \sum^L_{j=i}   e_ip_{i,j,k}  q_j  .
\end{align} 
Using $c_k$, (\ref{equARE2}) is rewritten as 
\begin{align}
 E_t =\sum^{L}_{k=1}  c_k (r_{k,k})^{t-1}.
\end{align} 
The conclusion then is proven. 
\end{IEEEproof}

This theorem shows  the relative approximation error  is represented as a linear combination of  exponential functions   $(\lambda_k)^t$ (where $k=1, \cdots, L$).  

From the relationship  between $F_t$ and $E_t$ and that between $R_t$ and $E_t$, we get the following corollaries.
\begin{corollary}
\label{cor1} 
The fitness value $F_t$ equals to
\begin{align}
F_t =f_{\opt}(1-\sum^{L}_{k=1}  c_k (\lambda_k)^{t-1}).
\end{align}
\end{corollary}

\begin{corollary}
\label{cor2}
The average convergence rate $R_t$ equals to
\begin{align}
R_t =1-\left(  \sum^{L}_{k=1}  c_k (\lambda_k)^{t-1} \frac{f_{\opt}}{ f_{\opt}-f_0}\right)^{1/t}.
\end{align}
\end{corollary}

In practice, the relative approximation error  is  calculated as follows:
\begin{algorithmic}[1]
\STATE given an initial probability distribution $\mathbf{p}_0$, the fitness error $\mathbf{e}$ and  matrix $\mathbf{R}$;

\STATE calculate  power factors $[p_{i,j,k}]$ where $i,j,k=1, \cdots, L$ using Definition~\ref{defMain};

\STATE  calculate coefficients $[c_k]$ (where $k=1, \cdots, L$) using (\ref{equAlpha});

\STATE calculate the relative approximation error  $E_t$ using Theorem~\ref{theMain}.

\end{algorithmic}

\section{Example (continued)}
\label{secExamplec}
This section applies Theorem~\ref{theMain} to the example in Section~\ref{secExample}.  The example is chosen for the sake of illustration.   Nevertheless Theorem~\ref{theMain} covers all $(1+1)$ strictly elitist EAs on any function under the unique condition.

We consider the OneMax function $
f_{\mathrm{one}}(x)$  first.  
The set $\{0, 1\}^4$ is split into $5$ subsets
\begin{align}
\mathcal{S}_i =\{x; |x|=i \}, && i=0,1, \cdots, 4.
\end{align} 
Each subset $\mathcal{S}_i$ is regarded as a state $i$. 

Transition probabilities $r_{i,j}=\Pr(\Phi_t\in \mathcal{S}_i \mid \Phi_{t-1} \in \mathcal{S}_j)$  are given by
\begin{equation}
r_{i,j}=\left\{
\begin{array}{lll}
\frac{j}{4}, & \mbox{if } j=i+1,\\
1-\frac{j}{4}, &\mbox{if } j=i,\\
0, &\mbox{otherwise}.
\end{array} 
\right.
\end{equation} 
Matrix $\mathbf{R}$  is  
\begin{align}
\label{equMatrix4}
\begin{pmatrix}
0.750 &0.500 &0.000 &0.000 \\
0.000 &0.500 &0.750 &0.000 \\
0.000 &0.000 &0.250 &1.000 \\
0.000 &0.000 &0.000 &0.000 \\ 
\end{pmatrix}
\end{align}

The fitness error  $e_i=i$ for $i=1, \cdots, 4$.
The fitness error 
vector is
\begin{align*} 
\mathbf{e}^T=(1, 2, 3, 4).
\end{align*}

Choose the initial probability distribution   in the non-optimal set to be
\begin{align*}
\mathbf{q}_0=(0, 0, 0, 1)^T.
\end{align*}

Using Definition~\ref{defMain}, we calculate matrix $[p_{i,j,k}]$ which is given by
\begin{align*}
[p_{1,j,k}]&=
\begin{pmatrix}
 0.750 &  1.500 &  2.250 &  3.000 \\ 
 0.000 &  -1.000 &  -3.000 &  -6.000 \\ 
 0.000 &  0.000 &  0.750 &  3.000 \\ 
 0.000 &  0.000 &  0.000 &  0.000 \\ 
\end{pmatrix},
\end{align*}
\begin{align*}
[p_{2,j,k}]&=
\begin{pmatrix}
 0.000 &  0.000 &  0.000 &  0.000 \\ 
 0.000 &  0.500 &  1.500 &  3.000 \\ 
 0.000 &  0.000 &  -0.750 &  -3.000 \\ 
 0.000 &  0.000 &  0.000 &  0.000 \\ 
\end{pmatrix},
\end{align*}
\begin{align*}
[p_{3,j,k}]&=
\begin{pmatrix}
 0.000 &  0.000 &  0.000 &  0.000 \\ 
 0.000 &  0.000 &  0.000 &  0.000 \\ 
 0.000 &  0.000 &  0.250 &  1.000 \\ 
 0.000 &  0.000 &  0.000 &  0.000 \\ 
\end{pmatrix},
\end{align*}
\begin{align*}
[p_{4,j,k}]&=
\begin{pmatrix}
 0.000 &  0.000 &  0.000 &  0.000 \\ 
 0.000 &  0.000 &  0.000 &  0.000 \\ 
 0.000 &  0.000 &  0.000 &  0.000 \\ 
 0.000 &  0.000 &  0.000 &  0.000 \\  
\end{pmatrix}.
\end{align*}

Using~(\ref{equAlpha}), we calculate coefficients $c_k$ (where $k=1, \cdots, 4$), given by
\begin{align*}
(0.750, 0.000, 0.000, 0.000 
) .
\end{align*}
Recall transition probabilities $r_{k,k}$ (where $ k=1, \cdots, 4)$ are
\begin{align*}
(0.750, 0.500, 0.250, 0.000).
\end{align*}
Using~(\ref{equARE3}), we calculate the relative approximation error  $ E_t$,  given by
 \begin{align}
 E_t =0.75^{t}.
 \end{align}

Furthermore, using Corollary~\ref{cor1}, we calculate the fitness value $F_t$, given by
 \begin{align}
 F_t =4(1- 0.75^{t}).
 \end{align}
And using Corollary~\ref{cor2}, we calculate the average convergence rate $R_t $, given by
 \begin{align}
 R_t =1-( 0.75^{t}   )^{1/t}=0.25.
 \end{align}
This means that  $E_t$ decays as fast as an exponential  function: $
E_t =0.75^t E_0$.

The analysis of  the quadratic  function $f_{\mathrm{squ}}(x)$  and logarithmic function  $f_{\mathrm{log}}(x)$ is almost the same as that of the OneMax function, except the fitness error vector $\mathbf{e}$. The results are summarised in Table~\ref{tab1}. Notice that the expressions for quadratic and logarithmic functions are more complex than that for the OneMax function.

\begin{table*}[htb]
\caption{Analytic expressions  of $F_t$, $E_t$ and $R_t$ in the example.}
\label{tab1}
\centering
\begin{tabular}{c|c }
\toprule
 function  &   $F_t$
\\ 
\midrule
  $f_{\mathrm{one}}=|x|$ &  $4\times(1-0.75\times0.75^{t-1})$ 
  \\ 
 $f_{\mathrm{squ}}=|x|^2$ &  $16\times(1-1.313\times0.75^{t-1}+0.375 \times0.5^{t-1})$   
\\ 
 $f_{\mathrm{log}}=\ln (|x|+1)$ &  $\ln(5)\times(1-0.416\times0.75^{t-1}-0.120 \times0.5^{t-1}-0.033\times0.25^{t-1})$   
\\
\midrule   &   $E_t$
\\ 
\midrule
  $f_{\mathrm{one}}=|x|$ &  $0.75\times0.75^{t-1}$ 
  \\ 
 $f_{\mathrm{squ}}=|x|^2$ &  $1.313\times0.75^{t-1}-0.375 \times0.5^{t-1}$   
\\ 
 $f_{\mathrm{log}}=\ln (|x|+1)$ &  $0.416\times0.75^{t-1}+0.120 \times0.5^{t-1}+0.033\times0.25^{t-1}$   
\\ 
\midrule
 &   $R_t$
\\ 
\midrule
  $f_{\mathrm{one}}=|x|$ &  $0.25$ 
  \\ 
 $f_{\mathrm{squ}}=|x|^2$ &  ${1-  (1.313 \times 0.75^{t-1}-0.375\times 0.5^{t-1}   )^{1/t}}$   
\\ 
 $f_{\mathrm{log}}=\ln (|x|+1)$ &  ${1-  (0.416\times0.75^{t-1}+0.120 \times0.5^{t-1}+0.033\times0.25^{t-1} )^{1/t}}$   
\\
\bottomrule
\end{tabular}
\end{table*}
  
\begin{figure}[ht]
\begin{center}
\begin{tikzpicture}
\begin{axis}[domain=0:35, width=7cm,height=5cm,
scaled ticks=false,
legend style={draw=none},legend pos= outer north east,
xlabel={$t$}, 
ylabel={$F_t$},  
]   
      \addplot [dotted] {4*(1-0.75*0.75^(x-1))};    
             \addplot [dashdotted ] {16*(1-1.313*0.75^(x-1)+0.375 *0.5^(x-1))}; 
                          \addplot [black,dashed ] {ln(5)*(1-0.416*0.75^(x-1)-0.120 *0.5^(x-1)-0.033*0.25^(x-1)))};    
                          \addplot table[only marks]   {F-f1-0.dat};  
  \addplot[mark=triangle*] table[ only marks]   {F-f2-0.dat};  
    \addplot[mark=square*] table[  only marks]   {F-f3-0.dat};  
         
    \legend{ $f_{\mathrm{one}}$, $f_{\mathrm{squ}}$, $f_{\mathrm{log}}$, $f^*_{\mathrm{one}}$, $f^*_{\mathrm{squ}}$, $f^*_{\mathrm{log}}$ }
        \end{axis}
\end{tikzpicture} 
\caption{Fitness value $F_t$.}
\label{fig-4}
\end{center}
\end{figure}
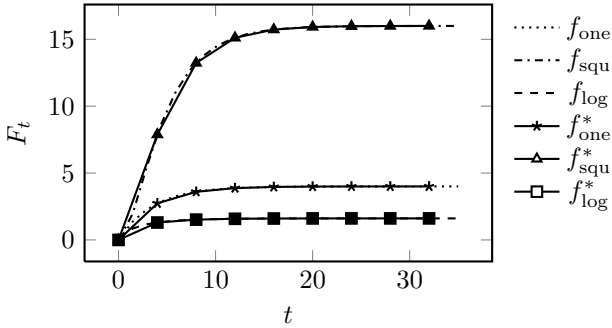

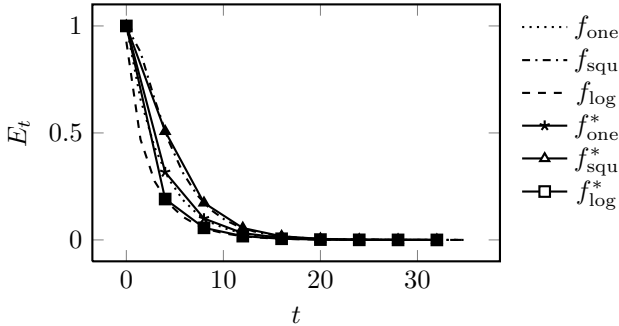
\begin{figure}[ht]
\begin{center}
\begin{tikzpicture}
\begin{axis}[domain=0:35, width=7cm,height=5cm,
scaled ticks=false,
legend style={draw=none},legend pos=outer north east,
xlabel={$t$}, 
ylabel={$E_t$}, 
]   
   
        \addplot [dotted] {0.75*0.75^(x-1)};   
             \addplot [dashdotted ] {1.313*0.75^(x-1)-0.375 *0.5^(x-1)};  
     \addplot [dashed ] {0.416*0.75^(x-1)+0.120 *0.5^(x-1)+0.033*0.25^(x-1)}; 
\addplot table[only marks]   {ARE-f1-0.dat};  
\addplot[mark=triangle*] table[ only marks]   {ARE-f2-0.dat};  
\addplot[mark=square*] table[  only marks]   {ARE-f3-0.dat};   
            
    \legend{$f_{\mathrm{one}}$, $f_{\mathrm{squ}}$, $f_{\mathrm{log}}$ , $f^*_{\mathrm{one}}$, $f^*_{\mathrm{squ}}$, $f^*_{\mathrm{log}}$ }
        \end{axis}
\end{tikzpicture} 
\caption{Relative approximation error  $E_t$.}
\label{fig-5}
\end{center}
\end{figure}

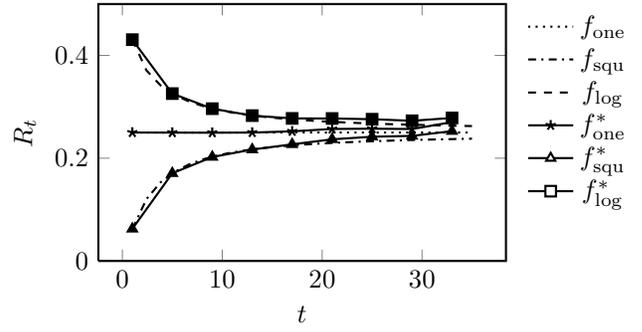
\begin{figure}[ht]
\begin{center}
\begin{tikzpicture}
\begin{axis}[domain=1:35, width=7cm,height=5cm,
scaled ticks=false,
legend style={draw=none},legend pos= outer north east,
xlabel={$t$}, 
ylabel={$R_t$}, 
ymin=0,
ymax=0.5,  
]    
      \addplot [dotted] {0.25};   
             \addplot [dashdotted ] {1-  (1.313 * 0.75^(x-1)-0.375* 0.5^(x-1)   )^(1/x)};   
                          \addplot [dashed ] {1-  (0.416*0.75^(x-1)+0.120 *0.5^(x-1)+0.033*0.25^(x-1) )^(1/x)};  
                           \addplot table[only marks]   {ACR-f1-0.dat};  
  \addplot[mark=triangle*] table[ only marks]   {ACR-f2-0.dat};  
    \addplot[mark=square*] table[  only marks]   {ACR-f3-0.dat}; 
    \legend{$f_{\mathrm{one}}$, $f_{\mathrm{squ}}$, $f_{\mathrm{log}}$, $f^*_{\mathrm{one}}$, $f^*_{\mathrm{squ}}$, $f^*_{\mathrm{log}}$ }
        \end{axis}
\end{tikzpicture} 
\caption{Average convergence rate $R_t$.}
\label{fig-6}
\end{center}
\end{figure}

Fig.~\ref{fig-4} demonstrates the fitness value $F_t$.  
Fig.~\ref{fig-5} presents the relative approximation error  $E_t$.   
Fig.~\ref{fig-6} illustrates the  average convergence rates  $R_t$. The theoretical predictions are consistent to the experimental results, labelled by $f^*$.

\section{Extension}
This section devotes to an extension    from (1+1) strictly elitist EAs to non-elitist or population-based EAs.

Many non-elitist or population-based EAs can be modelled by homogeneous Markov chains but    matrices $\mathbf{R}$   are not   upper triangular. Given  any matrix $\mathbf{R}$, according to  Schur's triangularisation theorem   (in textbook \cite[p508]{meyer2000matrix}),  there exists an upper triangular matrix $\tilde{\mathbf{R}}$ and unitary matrix $\mathbf{U}$ such that $\mathbf{R} =\mathbf{U} \tilde{\mathbf{R}} \mathbf{U}^*$. Then  the matrix iteration (\ref{equMatrixIteration}) can be rewritten as follows,
 \begin{align}
 \mathbf{q}_{t}&= \mathbf{R}^t  \mathbf{q}_0   = \mathbf{U} \tilde{\mathbf{R}}^t \mathbf{U}^* \mathbf{q}_0 . 
\label{equMatrixIteration2}  
\end{align}

Then the relative approximation error  equals to
\begin{align} 
 E_t 
 = \frac{      \mathbf{e}^T \mathbf{U} \tilde{\mathbf{R}}^t \mathbf{U}^* \mathbf{q}_0 }{f_{\opt}}  .
\end{align}  

Let $\tilde{\mathbf{e}}^T =\mathbf{e}^T \mathbf{U}$ and $\tilde{\mathbf{q}}_0= \mathbf{U}^* \mathbf{q}_0$, then the relative approximation error  can be rewritten as follows,
\begin{align} 
 E_t 
 = \frac{      \tilde{\mathbf{e}}^T \tilde{\mathbf{R}}^t \tilde{ \mathbf{q}}_0 }{f_{\opt}}.
\end{align} 
Since $\tilde{\mathbf{R}}$ is an upper triangular matrix, the analysis of $E_t$ becomes the problem of expressing the matrix power $\tilde{\mathbf{R}}^t$. If $\tilde{\mathbf{R}}$ is an upper triangular matrix with unique diagonal entries, Theorem~\ref{theMain} can be applied directly. If this does not hold,  a similar analysis can be conduced (but in a separate paper). Therefore, in theory it is feasible to apply the approach to non-elitist or population-based EAs too.

Furthermore, even if  exact transition probabilities   are unknown, it is still possible to apply the method  to bounding the relative approximation error. The idea is simple. We construct an upper triangular matrix  $\mathbf{S}=[s_{i,j}]$  so that the matrix iteration using $\mathbf{S}$ is slower than that using $\mathbf{S}$. That is
 $\mathbf{e}^T \mathbf{S}^t \mathbf{q}_0 \ge \mathbf{e}^T \mathbf{R}^t \mathbf{q}_0.$ For example,  the simplest matrix $\mathbf{S}$ is
\begin{equation}
s_{i,j}=\left\{
\begin{array}{lll}
1-r_{j,j}, & \mbox{if } j=i+1,\\
r_{i,i}, &\mbox{if } j=i,\\
0, &\mbox{otherwise}.
\end{array} 
\right.
\end{equation} 
  This issue will be discussed  in a separate paper.

\section{Conclusions}
\label{secConclusions} 
In this paper, the solution quality  of  an EA is measured by the relative approximation error,  that is
\begin{equation}
E_t= 1-\frac{F_t}{f_{\opt}}.
\end{equation}
Then   an analytic expression of the relative approximation error $E_t$ is presented for any (1+1) strictly elitist EAs on any fitness function. Provided that  transition probabilities $r_{i,i} \neq r_{j,j}$ for any $i \neq j$, the formula is given by
  \begin{align}  
 E_t =\sum^{L}_{k=1}  c_k \lambda_k^{t-1},
\end{align}
where $\lambda_k=r_{k,k}$ (where $k=1, \cdots, L$) are eigenvalues of transition submatrix $\mathbf{R}$ and  $c_k$ are coefficients.  

The above formula is also useful to fixed budget analysis.  Since the exact expression of the fitness value $F_t$ is  
\begin{align}
F_t =f_{\opt}\left( 1-\sum^{L}_{k=1}  c_k (\lambda_k)^{t-1}\right),
\end{align} 
a good bound on $F_t$  should be represented in the form of a combination of exponential functions of $t$.  

The work is a further  development of  the  average convergence rate~\cite{he2016average}. The exact expression of the  average convergence rate $R_t$ is  
\begin{align}
R_t =1-\left(  \sum^{L}_{k=1}  c_k (\lambda_k)^{t-1} \frac{f_{\opt}}{ f_{\opt}-f_0}\right)^{1/t}.
\end{align}

The approach is  promising.  Using Schur's triangularization theorem,  it is feasible to make a similar analysis for non-elitist or population-based EAs if they are modelled by homogeneous  Markov chains.  

Our next work is to present a closed form for  (1+1) strictly elitist EAs whose transition matrices are upper triangular  but diagonal entries are not unique.

\subsection*{Acknowledgement:}  The work was supported by  the EPSRC under Grant EP/I009809/1.
 


\begin{thebibliography}{10}
\providecommand{\url}[1]{#1}
\csname url@samestyle\endcsname
\providecommand{\newblock}{\relax}
\providecommand{\bibinfo}[2]{#2}
\providecommand{\BIBentrySTDinterwordspacing}{\spaceskip=0pt\relax}
\providecommand{\BIBentryALTinterwordstretchfactor}{4}
\providecommand{\BIBentryALTinterwordspacing}{\spaceskip=\fontdimen2\font plus
\BIBentryALTinterwordstretchfactor\fontdimen3\font minus
  \fontdimen4\font\relax}
\providecommand{\BIBforeignlanguage}[2]{{%
\expandafter\ifx\csname l@#1\endcsname\relax
\typeout{** WARNING: IEEEtran.bst: No hyphenation pattern has been}%
\typeout{** loaded for the language `#1'. Using the pattern for}%
\typeout{** the default language instead.}%
\else
\language=\csname l@#1\endcsname
\fi
#2}}
\providecommand{\BIBdecl}{\relax}
\BIBdecl

\bibitem{oliveto2009analysis}
P.~S. Oliveto, J.~He, and X.~Yao, ``Analysis of the (1+1)-{EA} for finding
  approximate solutions to vertex cover problems,'' \emph{IEEE Transactions on
  Evolutionary Computation}, vol.~13, no.~5, pp. 1006 --1029, 2009.

\bibitem{friedrich2010approximating}
T.~Friedrich, J.~He, N.~Hebbinghaus, F.~Neumann, and C.~Witt, ``Approximating
  covering problems by randomized search heuristics using multi-objective
  models,'' \emph{Evolutionary Computation}, vol.~18, no.~4, pp. 617--633,
  2010.

\bibitem{witt2005worst}
C.~Witt, ``Worst-case and average-case approximations by simple randomized
  search heuristics,'' in \emph{Proceedings of the 22nd Annual Conference on
  Theoretical Aspects of Computer Science}.\hskip 1em plus 0.5em minus
  0.4em\relax Springer-Verlag, 2005, pp. 44--56.

\bibitem{yu2012approximation}
Y.~Yu, X.~Yao, and Z.-H. Zhou, ``On the approximation ability of evolutionary
  optimization with application to minimum set cover,'' \emph{Artificial
  Intelligence}, no. 180-181, pp. 20--33, 2012.

\bibitem{lai2014performance}
X.~Lai, Y.~Zhou, J.~He, and J.~Zhang, ``Performance analysis of evolutionary
  algorithms for the minimum label spanning tree problem,'' \emph{IEEE
  Transactions on Evolutionary Computation}, vol.~18, no.~6, pp. 860--872,
  2014.

\bibitem{he2003analysis}
J.~He and X.~Yao, ``An analysis of evolutionary algorithms for finding
  approximation solutions to hard optimisation problems,'' in \emph{Proceedings
  of IEEE 2003 Congress on Evolutionary Computation}.\hskip 1em plus 0.5em
  minus 0.4em\relax IEEE Press, 2003, pp. 2004--2010.

\bibitem{suzuki1995markov}
J.~Suzuki, ``A {Markov} chain analysis on simple genetic algorithms,''
  \emph{IEEE Transactions on Systems, Man and Cybernetics}, vol.~25, no.~4, pp.
  655--659, 1995.

\bibitem{suzuki1998further}
------, ``A further result on the markov chain model of genetic algorithms and
  its application to a simulated annealing-like strategy,'' \emph{IEEE
  Transactions on Systems, Man, and Cybernetics, Part B: Cybernetics}, vol.~28,
  no.~1, pp. 95--102, 1998.

\bibitem{he1999convergence}
J.~He and L.~Kang, ``On the convergence rate of genetic algorithms,''
  \emph{Theoretical Computer Science}, vol. 229, no. 1-2, pp. 23--39, 1999.

\bibitem{rudolph1997local}
G.~Rudolph, ``Local convergence rates of simple evolutionary algorithms with
  {Cauchy} mutations,'' \emph{IEEE Transactions on Evolutionary Computation},
  vol.~1, no.~4, pp. 249--258, 1997.

\bibitem{rudolph1997convergence}
------, ``Convergence rates of evolutionary algorithms for a class of convex
  objective functions,'' \emph{Control and Cybernetics}, vol.~26, pp. 375--390,
  1997.

\bibitem{rudolph2013convergence}
------, ``Convergence rates of evolutionary algorithms for quadratic convex
  functions with rank-deficient hessian,'' in \emph{Adaptive and Natural
  Computing Algorithms}.\hskip 1em plus 0.5em minus 0.4em\relax Springer, 2013,
  pp. 151--160.

\bibitem{ming2006convergence}
L.~Ming, Y.~Wang, and Y.-M. Cheung, ``On convergence rate of a class of genetic
  algorithms,'' in \emph{Proceedings of 2006 World Automation Congress}.\hskip
  1em plus 0.5em minus 0.4em\relax IEEE, 2006, pp. 1--6.

\bibitem{schmitt2001importance}
F.~Schmitt and F.~Rothlauf, ``On the importance of the second largest
  eigenvalue on the convergence rate of genetic algorithms,'' in
  \emph{Proceedings of 2001 Genetic and Evolutionary Computation Conference},
  H.~Beyer, E.~Cantu-Paz, D.~Goldberg, Parmee, L.~Spector, and D.~Whitley,
  Eds.\hskip 1em plus 0.5em minus 0.4em\relax Morgan Kaufmann Publishers, 2001,
  pp. 559--564.

\bibitem{ding2001convergence}
L.~Ding and L.~Kang, ``Convergence rates for a class of evolutionary algorithms
  with elitist strategy,'' \emph{Acta Mathematica Scientia}, vol.~21, no.~4,
  pp. 531--540, 2001.

\bibitem{jansen2012fixed}
T.~Jansen and C.~Zarges, ``Fixed budget computations: A different perspective
  on run time analysis,'' in \emph{Proceedings of the 14th Annual Conference on
  Genetic and Evolutionary Computation}.\hskip 1em plus 0.5em minus 0.4em\relax
  ACM, 2012, pp. 1325--1332.

\bibitem{jansen2014performance}
------, ``Performance analysis of randomised search heuristics operating with a
  fixed budget,'' \emph{Theoretical Computer Science}, vol. 545, pp. 39--58,
  2014.

\bibitem{he2016average}
J.~He and G.~Lin, ``Average convergence rate of evolutionary algorithms,''
  \emph{IEEE Transactions on Evolutionary Computation}, vol.~20, no.~2, pp.
  316--321, 2016.

\bibitem{he2015easiest}
J.~He, T.~Chen, and X.~Yao, ``On the easiest and hardest fitness functions,''
  \emph{IEEE Transactions on Evolutionary Computation}, vol.~19, no.~2, pp.
  295--305, 2015.

\bibitem{he2003towards}
J.~He and X.~Yao, ``Towards an analytic framework for analysing the computation
  time of evolutionary algorithms,'' \emph{Artificial Intelligence}, vol. 145,
  no. 1-2, pp. 59--97, 2003.

\bibitem{rudolph1998finite}
G.~Rudolph, ``Finite {M}arkov chain results in evolutionary computation: a tour
  d'horizon,'' \emph{Fundamenta Informaticae}, vol.~35, no.~1, pp. 67--89,
  1998.

\bibitem{shur2011simple}
W.~Shur, ``A simple closed form for triangular matrix powers,''
  \emph{Electronic Journal of Linear Algebra}, vol.~22, pp. 1000--1003, 2011.

\bibitem{huang1978efficient}
C.~Huang, ``An efficient algorithm for computing powers of triangular
  matrices,'' in \emph{Proceedings of the 1978 ACM Annual Conference-Volume
  2}.\hskip 1em plus 0.5em minus 0.4em\relax ACM, 1978, pp. 954--957.

\bibitem{meyer2000matrix}
C.~Meyer, \emph{Matrix Analysis and Applied Linear Algebra}.\hskip 1em plus
  0.5em minus 0.4em\relax SIAM, 2000.

\end{thebibliography}

\end{document}